\documentclass{article}
\usepackage{arxiv}

\usepackage[utf8]{inputenc} 
\usepackage[T1]{fontenc}    
\usepackage{hyperref}       
\usepackage{url}            
\usepackage{booktabs}       
\usepackage{amsfonts}       
\usepackage{nicefrac}       
\usepackage{microtype}      
\usepackage{lipsum}
\usepackage{graphicx}
\graphicspath{ {./images/} }

\title{A Study of Rule Omission in Raven’s Progressive Matrices}

\author{
 Binze Li \\
  University of California, Los Angeles \\
  Los Angeles, CA 90024 \\
  \texttt{binzeli@g.ucla.edu} \\
}

\begin{document}
\maketitle
\begin{abstract}
Analogical reasoning lies at the core of human cognition and remains a fundamental challenge for artificial intelligence. Raven’s Progressive Matrices (RPM) serve as a widely used benchmark to assess abstract reasoning by requiring the inference of underlying structural rules. While many vision-based and language-based models have achieved success on RPM tasks, it remains unclear whether their performance reflects genuine reasoning ability or reliance on statistical shortcuts. This study investigates the generalization capacity of modern AI systems under conditions of incomplete training by deliberately omitting several structural rules during training. Both sequence-to-sequence transformer models and vision-based architectures such as CoPINet and the Dual-Contrast Network are evaluated on the Impartial-RAVEN (I-RAVEN) dataset. Experiments reveal that although transformers demonstrate strong performance on familiar rules, their accuracy declines sharply when faced with novel or omitted rules. Moreover, the gap between token-level accuracy and complete answer accuracy highlights fundamental limitations in current approaches. These findings provide new insights into the reasoning mechanisms underlying deep learning models and underscore the need for architectures that move beyond pattern recognition toward robust abstract reasoning.
\end{abstract}


\section{Introduction}
Analogical reasoning is a fundamental aspect of human cognition, enabling individuals to recognize, apply, and manipulate abstract patterns across varying contexts. One of the most widely recognized benchmarks for assessing this ability is Raven’s Progressive Matrices (RPM), a non-verbal test designed to measure fluid intelligence through the completion of complex visual patterns \cite{zhang2019raven}. This task requires the identification of underlying relational structures governing a given set of images, challenging both humans and artificial intelligence (AI) models to engage in abstract reasoning. 

In recent years, deep learning models—both vision-based and language-based—have been increasingly evaluated on RPM tasks to assess their capacity for relational and analogical reasoning. While previous studies have primarily focused on training models to learn the complete set of rules governing the RPM dataset, a critical question remains unanswered: how well do these models generalize when their training is incomplete? Specifically, if certain reasoning rules are omitted during the training phase, can these models still demonstrate an ability to generalize and infer solutions involving the missing rules? 

This study investigates the impact of selectively removing one or two rules during training and subsequently testing models on tasks involving the omitted rules. By systematically examining this effect, the goal is to provide new insights into the generalization abilities of contemporary AI architectures. A deeper understanding of these mechanisms is essential in determining whether current machine learning models exhibit true analogical reasoning or are merely leveraging statistical correlations present in the dataset. 

Furthermore, previous works have studied the effectiveness of various architectures, including the use of Vision Transformer Contrastive Networks (ViTCN) \cite{song2021vitcn}, contrastive perceptual-conceptual models \cite{yang2020cognitive}, and neuro-symbolic approaches \cite{hersche2021neurovector}. While these models have achieved considerable success in solving RPM tasks, questions about their generalization abilities remain open. More recently, research in natural language processing has explored in-context analogical reasoning with large language models, but they have limitations in grasping spatial and relational reasoning tasks 
\cite{brown2020lm, wei2022cot, lake2017bbls, webb2023language}.

This study conducts extensive testing on both vision-based and language-based models to evaluate their capacity for analogical reasoning in the context of RPM tasks. Beyond examining language models trained from scratch, the evaluation also includes two representative vision-based models to provide a comparative perspective across modalities. By selectively removing rules during training and assessing model performance on those omitted rules during testing, the study delivers a systematic analysis of generalization abilities. The findings highlight differences in how vision-based and language-based systems acquire and apply abstract rules, offering new insights into the underlying reasoning mechanisms that shape model performance on relational and analogical tasks.

\section{Background}

\subsection{Raven’s Progressive Matrices}
Raven’s Progressive Matrices (RPM) were originally developed to measure fluid intelligence by presenting individuals with a series of abstract visual patterns, requiring them to infer the missing component based on underlying relationships. The test is widely used in cognitive science and artificial intelligence research as a benchmark for evaluating abstract reasoning abilities. The RAVEN image dataset \cite{zhang2019raven} extends the traditional RPM by incorporating explicit structural representations and multiple rule types, making it a crucial dataset for machine learning research in relational and analogical reasoning. 

In the standard RPM setup, each problem is presented as a 3×3 matrix of images. The panels in the first two rows follow consistent underlying relational rules, and the challenge is to infer these rules in order to complete the third row. The final panel in the third row is left blank, and the task is to determine the correct missing panel by applying the inferred rule and selecting the answer from a set of eight candidate choices.

\subsection{Computational Models}
Several machine learning approaches have been proposed to tackle RPM, with varying levels of success. Vision-based models, such as the Vision Transformer Contrastive Network (ViTCN) developed by Song et al., employ self-attention mechanisms to effectively capture complex relational dependencies within the images. Similarly, the Dual-Contrast Network \cite{zhuo2020dcn} and contrastive perceptual-conceptual processing models \cite{yang2020cognitive} introduce cognitive science-inspired methodologies to improve abstract reasoning performance. These architectures attempt to bridge the gap between human-like cognitive reasoning and machine learning's ability to process large amounts of structured data. 

Another promising direction involves neuro-symbolic approaches, such as the Neuro-vector-symbolic architecture \cite{hersche2021neurovector}, which integrate neural network learning with symbolic reasoning processes. These methods aim to enhance the interpretability and logical consistency of AI systems when solving abstract reasoning tasks. The ability of neuro-symbolic models to integrate learned representations with structured reasoning frameworks makes them an interesting area of study for tasks involving high-level cognition. 

Beyond vision-based models, language-based models have also been examined in the context of RPM tasks. Hu et al. explored in-context analogical reasoning using pre-trained language models, analyzing their ability to infer relational structures from textual descriptions rather than direct visual input \cite{hu2022analogical}. This line of research suggests that language models may enhance the ability to recognize and reason about abstract patterns. Recent work in natural language processing has increasingly examined the capacity of large language models (LLMs) to perform analogical reasoning through in-context learning and few-shot learning techniques \cite{brown2020lm}. These approaches reveal that pretrained transformers can generalize to new reasoning tasks with minimal supervision, highlighting their potential for flexible adaptation \cite{brown2020lm}. Further studies have shown that prompting LLMs with explicit reasoning chains can substantially improve their performance on logical and mathematical analogies, indicating that structured reasoning cues enable models to better exploit latent reasoning abilities \cite{wei2022cot}. Despite these advances, purely text-based representations remain limited, as true abstract reasoning requires sensitivity to causal, spatial, and relational structures that are only partially captured in language \cite{lake2017bbls}. More recent findings suggest that while LLMs can demonstrate emergent analogical reasoning capabilities, their reasoning processes differ from human cognition and show notable weaknesses when applied to visual or relational domains \cite{ webb2023language}.

\section{Experiments}
\subsection{Dataset}
This work uses the Impartial-RAVEN (I-RAVEN) dataset instead of the original RAVEN dataset \cite{hu2019iraven}. Although RAVEN has been widely adopted for evaluating abstract reasoning, its answer-set construction contains severe defects, such as annotation biases and spurious correlations that allow models to exploit unintended shortcuts. For example, in RAVEN, the incorrect answer candidates are often generated in a way that simultaneously violates multiple rules or introduces imbalanced attribute distributions, which makes the correct answer stand out as the “most similar” option rather than requiring proper reasoning over the underlying matrix rules. These issues mean that a model can achieve high performance by relying on surface-level statistical cues rather than demonstrating genuine relational and analogical reasoning. To overcome these shortcomings, I-RAVEN employs the Attribute Bisection Tree (ABT) algorithm to generate answer sets where distractors differ from the ground truth in a balanced and systematic manner, ensuring that no option is unfairly distinguished by irrelevant biases. This correction removes misleading cues, enforces impartiality across the answer set, and compels models to infer the true structural relationships within the matrix. 

The I-RAVEN dataset is structured around five rule-governing attributes: Number, Position, Type, Size, and Color. Each attribute is governed by one of four possible rules: Constant, Progression, Arithmetic, and Distribute Three. These rules define how the attributes change or remain consistent across the matrix:
\begin{itemize}
  \item \textbf{Constant}: The attribute remains unchanged across the rows or columns of the matrix.
  \item \textbf{Progression}: The attribute changes in a consistent manner (e.g., increasing or decreasing) across the rows or columns.
  \item \textbf{Arithmetic}: The attribute follows a mathematical relationship (e.g., addition or subtraction) between elements in the matrix.
  \item \textbf{Distribute Three}: The attribute distributes three distinct values across the elements in the matrix.
\end{itemize}

To evaluate the generalization capabilities of models under incomplete training conditions, two modified versions of the I-RAVEN dataset are created by selectively removing specific rules during training:

\begin{itemize}
  \item \textbf{Remove 1 Rule}: In this subset, we removed all instances where any of the four attributes follows the \textit{Progression} rule.
  \item \textbf{Remove 2 Rules}: In this subset, we removed all instances where any of the four attributes follows the \textit{Progression} and \textit{Arithmetic} rules.
\end{itemize}

For the sequence-to-sequence transformer model, the visual data are transformed into a textual representation to facilitate processing. Each image is converted into a structured text format describing the attributes of the shapes. The attributes and their possible values are as follows:

\begin{itemize}
  \item \textbf{Type}: The shape type, which can be one of \texttt{["none", "triangle", "square", "pentagon", "hexagon", "circle"]}.
  \item \textbf{Size}: The size of the shape, which takes values from \texttt{[0.4, 0.5, 0.6, 0.7, 0.8, 0.9]}.
  \item \textbf{Color}: The color intensity of the shape, represented by values from \texttt{[255, 224, 196, 168, 140, 112, 84, 56, 28, 0]}.
  \item \textbf{Angle}: The orientation of the shape, which can be one of \texttt{[-135, -90, -45, 0, 45, 90, 135, 180]}.
\end{itemize}

For example, the text representation \texttt{[0.5, 0.5, 1, 1], 3, 3, 5, 7} corresponds to a shape centered, a pentagon (\texttt{Type = 3}), a size value of 0.7 (index 3 in \texttt{Size}), a color value of 112 (index 5 in \texttt{Color}), and an angle of 180 degrees (index 7 in \texttt{Angle}). This transformation allows the sequence-to-sequence transformer to process the RPM tasks as a sequence prediction problem.

\subsection{Models}
This study evaluates three models: a sequence-to-sequence transformer model and two existing vision-based models.

A sequence-to-sequence transformer model (seq-seq) is applied to process the textual representations of the RAVEN dataset. This model was inspired by the framework from the work on human-like systematic generalization through meta-learning \cite{lake2023systematic}. The transformer architecture is chosen for its ability to capture long-range dependencies and relational structures, which are critical for solving RPM tasks.The model consists of an encoder-decoder architecture with multi-head self-attention mechanisms. The encoder processes the input sequence (e.g., textual descriptions of the matrix elements), while the decoder generates the predicted output sequence (e.g., the missing element in the matrix). The model was trained from scratch using a cross-entropy loss function to minimize the discrepancy between the predicted and actual sequences.

CoPINet, a vision-based model proposed was designed for perceptual inference through contrastive learning \cite{zhang2019copinet}. CoPINet employs a dual-path architecture that contrasts perceptual and conceptual representations to improve reasoning performance. The model was implemented using the original code provided in the authors' GitHub repository. It was trained and evaluated on the modified subsets.
The second vision-based model evaluated was the Dual-Contrast Network (dcnet) \cite{zhuo2020dcn}. This model uses a dual-contrast mechanism to enhance abstract reasoning by comparing both low-level perceptual features and high-level conceptual features. Similar to CoPINet, publicly available implementation from the authors' GitHub repository was used. It was trained and evaluated on the modified subsets.

\begin{table}[ht]
\centering
\caption{Performance metrics on the ``Seq remove 1 Different rule'' setting across various configurations}
\begin{tabular}{lccccccc}
\toprule
 & \textbf{Center} & \textbf{2x2} & \textbf{3x3} & \textbf{O - IC} & \textbf{O - IG} & \textbf{L - R} & \textbf{U - D} \\
\midrule
\textbf{Accuracy (by token)} & 96.88\% & 71.88\% & 66.32\% & 88.21\% & 74.94\% & 86.91\% & 87.02\% \\
\textbf{Accuracy (correct choice)} & 45.31\% & 49.76\% & 44.71\% & 45.91\% & 46.03\% & 49.28\% & 47.96\% \\
\textbf{F1} & 0.869 & 0.716 & 0.653 & 0.8815 & 0.747 & 0.8702 & 0.8704 \\
\textbf{TER} & 0.131 & 0.281 & 0.337 & 0.118 & 0.251 & 0.1309 & 0.1298 \\
\bottomrule
\end{tabular}
\label{tab:seq_remove_1}
\end{table}

\begin{table}[ht]
\centering
\caption{Performance metrics on the ``Seq remove 1 Same rule'' setting across various configurations}
\begin{tabular}{lccccccc}
\toprule
 & \textbf{Center} & \textbf{2x2} & \textbf{3x3} & \textbf{O - IC} & \textbf{O - IG} & \textbf{L - R} & \textbf{U - D} \\
\midrule
\textbf{Accuracy (by token)} & 99.97\% & 80.8\% & 79.21\% & 99.9\% & 84.46\% & 100\% & 99.9\% \\
\textbf{Accuracy (correct choice)} & 100\% & 82.45\% & 79.69\% & 100\% & 86.18\% & 100\% & 100\% \\
\textbf{F1} & 0.9996 & 0.804 & 0.784 & 0.999 & 0.843 & 1 & 0.999 \\
\textbf{TER} & 0.0003 & 0.192 & 0.208 & 0 & 0.155 & 0 & 0 \\
\bottomrule
\end{tabular}
\label{tab:seq_remove_1_same}
\end{table}

\section{Results}
The experimental results reveal several key patterns in model performance across different rule-removal scenarios. The analysis focuses on two primary dimensions: rule familiarity (whether testing rules appeared in training) and task complexity (number of rules removed). The \textbf{“Different Rule”} setting refers to scenarios in which the training dataset omits one or two rules, and the model is then tested on data containing only the omitted rule. In contrast, the \textbf{“Same Rule”} setting refers to scenarios where the training dataset includes specific rules, and the testing dataset is restricted to those same rules, ensuring full overlap between training and testing conditions.

\paragraph{Rule Familiarity Effects} Tables 1-4 demonstrate a consistent performance gap between "Same" and "Different" rule conditions for the seq-seq transformer. Two complementary accuracy metrics provided detailed insights: 1) Accuracy (by token) - Measuring the percentage of individual tokens predicted correctly and 2) Accuracy (correct choice) - Assessing the percentage of completely correct answers selected from multiple choices. For instance, Table \ref{tab:seq_remove_1} shows that in the "Seq remove 1 Different rule" setting, token-level accuracy ranges from 66.32\% (3x3) to 96.88\% (Center), while Table \ref{tab:seq_remove_1_same} reveals better performance for all configurations for the corresponding "Same rule" condition. This pattern repeats in the two-rule removal scenario (Tables \ref{tab:seq_remove_2}-\ref{tab:seq_remove_2_same}), with performance differences exceeding 15 percentage points across all configurations.

\begin{figure}[ht]
\centering
\includegraphics[width=\linewidth, keepaspectratio]{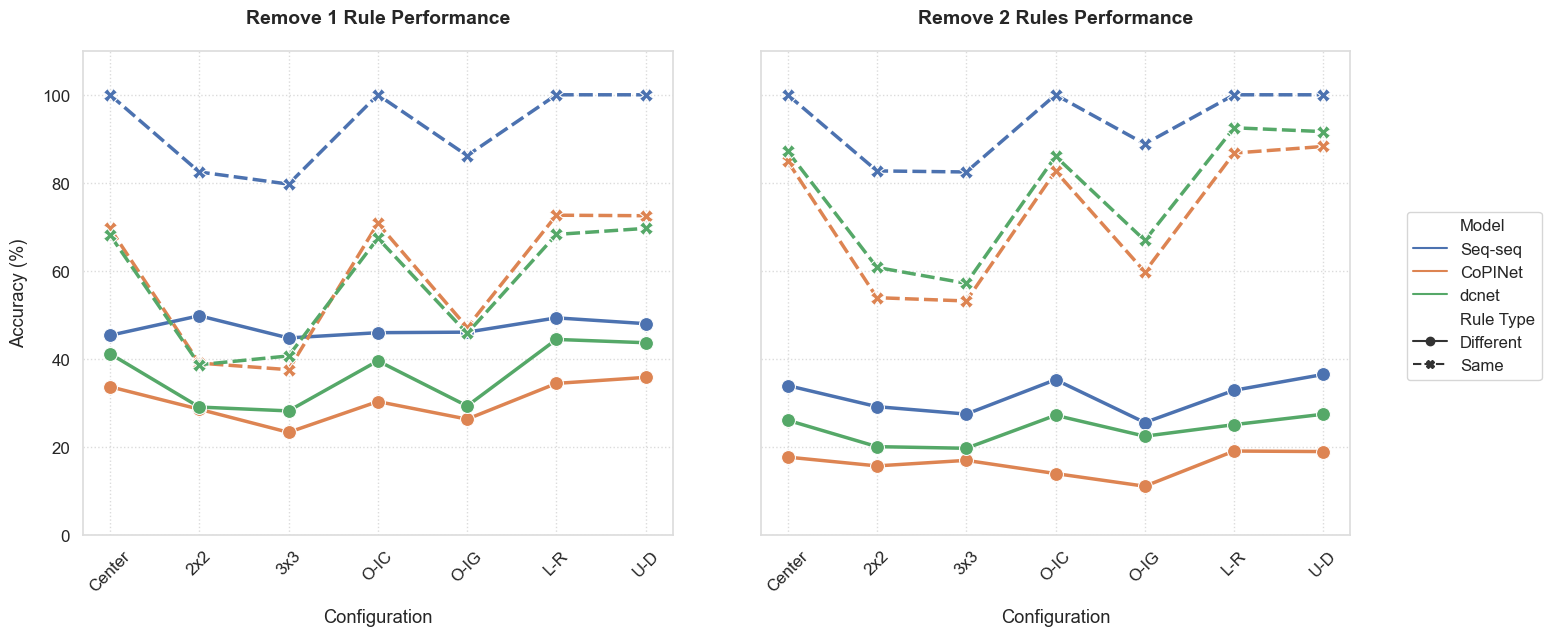}
\caption[Short caption for LoF]{Model Performance Comparison Across Rule Removal Scenarios}
\label{fig:fig1}
\end{figure}

\paragraph{Task Complexity Impact}
Figure \ref{fig:fig1} and Table \ref{tab:average_performance} clearly illustrate that removing one rule (Tables 1-2) yields substantially better results than removing two rules (Tables 3-4). The average accuracy drops from 47.00\% to 31.47\% for Seq-seq in "Different" conditions (Table 5), while maintaining high performance (92.62\% to 93.42\%) in "Same" conditions. This complexity effect is further evidenced by the TER (Token Error Rate) metrics in Tables 1-4, which show error rates doubling in some cases when moving from one- to two-rule removal.

\paragraph{Configuration Performance}
Spatial configuration significantly impacts results, as shown in Tables 1-4. Center and O-IC configurations consistently outperform others, with Center achieving 96.88\% token accuracy in Table \ref{tab:seq_remove_1} and O-IC achieving 84.46\% token accuracy. In contrast, 2x2 and 3x3 configurations show the weakest performance across all scenarios, suggesting these complex arrangements hinder rule learning.

\paragraph{Model Comparison}
Table \ref{tab:average_performance} provides a comprehensive comparison of model performance. Seq-seq demonstrates clear superiority, particularly in "Same" conditions where it achieves >90\% accuracy versus <80\% for the two vision models. This advantage persists in "Different" conditions where the two vision models perform lower than 30\% accuracy on two-rules-removal scenarios.

\begin{table}[ht]
\centering
\caption{Performance metrics on the ``Seq remove 2 Different rule'' setting across various configurations}
\begin{tabular}{lccccccc}
\toprule
 & \textbf{Center} & \textbf{2x2} & \textbf{3x3} & \textbf{O - IC} & \textbf{O - IG} & \textbf{L - R} & \textbf{U - D} \\
\midrule
\textbf{Accuracy (by token)} & 82.03\% & 51.51\% & 51.08\% & 84.62\% & 59.05\% & 81.87\% & 82.24\% \\
\textbf{Accuracy (correct choice)} & 33.89\% & 29.09\% & 27.4\% & 35.22\% & 25.48\% & 32.81\% & 36.42\% \\
\textbf{F1} & 0.819 & 0.498 & 0.495 & 0.848 & 0.579 & 0.819 & 0.821 \\
\textbf{TER} & 0.18 & 0.485 & 0.489 & 0.154 & 0.409 & 0.181 & 0.178 \\
\bottomrule
\end{tabular}
\label{tab:seq_remove_2}
\end{table}

\begin{table}[ht]
\centering
\caption{Performance metrics on the ``Seq remove 2 Same rule'' setting across various configurations}
\begin{tabular}{lccccccc}
\toprule
 & \textbf{Center} & \textbf{2x2} & \textbf{3x3} & \textbf{O - IC} & \textbf{O - IG} & \textbf{L - R} & \textbf{U - D} \\
\midrule
\textbf{Accuracy (by token)} & 100\% & 85.46\% & 84.79\% & 100\% & 87.93\% & 100\% & 100\% \\
\textbf{Accuracy (correct choice)} & 100\% & 82.69\% & 82.45\% & 100\% & 88.82\% & 100\% & 100\% \\
\textbf{F1} & 1 & 0.854 & 0.847 & 1 & 0.879 & 1 & 1 \\
\textbf{TER} & 0 & 0.145 & 0.152 & 0 & 0.121 & 0 & 0 \\
\bottomrule
\end{tabular}
\label{tab:seq_remove_2_same}
\end{table}

\begin{table}[ht]
\centering
\caption{Average Accuracy (\%) Across Rule Removal Scenarios}
\begin{tabular}{lcccc}
\toprule
 & \multicolumn{2}{c}{\textbf{Remove 1 Rule}} & \multicolumn{2}{c}{\textbf{Remove 2 Rules}} \\
\cmidrule(lr){2-3} \cmidrule(lr){4-5}
\textbf{Model} & \textbf{Different} & \textbf{Same} & \textbf{Different} & \textbf{Same} \\
\midrule
Seq-seq & \textbf{47.00} & \textbf{92.62} & \textbf{31.47} & \textbf{93.42} \\
CoPINet & 30.29 & 58.45 & 16.13 & 72.75 \\
dcnet & 36.43 & 56.91 & 23.93 & 77.43 \\
\bottomrule
\end{tabular}
\label{tab:average_performance}
\end{table}

\section{Discussion}
\paragraph{Limitations in Generalization Capabilities}
The study reveals a fundamental limitation in current AI systems' ability to generalize learned rules to novel configurations. While models demonstrate strong performance on familiar patterns seen during training, their reasoning capabilities break down when faced with novel rule combinations. This performance gap suggests that current approaches rely more on pattern memorization than on developing genuine abstract reasoning skills. The transformer architecture exemplifies this limitation, achieving near-perfect accuracy on trained rules but showing significant degradation when encountering unfamiliar configurations. This behavior becomes particularly pronounced in complex scenarios with multiple interacting rules, where models exhibit substantially poorer performance compared to human reasoners facing similar challenges.

\paragraph{The Illusion of High Token-Level Accuracy}
The apparent success reflected in token-level accuracy metrics presents a misleading picture of true reasoning capabilities. While models achieve relatively high scores in predicting individual tokens (e.g., 96.88\% for Center configuration in one-rule removal), this performance fails to translate to correct final answers (only 45.31\% accuracy). This discrepancy occurs because the answer choices in RPM tasks often differ by just one or two tokens - meaning a model that misses even a single critical token will select the entirely wrong answer, despite being mostly correct. This phenomenon explains why the gap between token-level accuracy (measuring partial correctness) and correct choice accuracy (measuring complete solutions) is so dramatic. The models demonstrate "fragile correctness" - they can solve most of the problem correctly but fail at the final hurdle of producing a fully accurate solution.

\paragraph{Performance Improvement in Vision Models}
Contrary to initial expectations that accuracy should remain comparable across "Remove 1 Same" and "Remove 2 Same" conditions (since both test on rules seen during training), the vision models actually demonstrate superior performance in the latter scenario. This counterintuitive result suggests that the additional rule removal during training may have created a more favorable learning environment by reducing the complexity and variability of the remaining rule combinations. When more rules are removed, the leftover complexity decreases. The effect appears particularly pronounced in vision-based models, potentially due to their greater sensitivity to pattern complexity compared to sequence-based approaches.

\paragraph{Architectural and Representational Considerations}
Our results demonstrate that architectural choices significantly impact reasoning performance, with simpler spatial configurations proving substantially more learnable than complex arrangements. This finding suggests current attention mechanisms may struggle with distributed relational patterns, performing better when rules can be localized to specific regions. Interestingly, these constraints mirror known limitations in human cognitive processing, revealing potential parallels between biological and artificial reasoning boundaries. The superior performance of sequence-based approaches over vision-based alternatives raises important questions about modality effects, pointing to the value of intermediate representations that bridge perceptual and symbolic reasoning. These representational considerations may hold the key to developing more robust reasoning systems.

\section{Conclusion}
This study provides a systematic analysis of how contemporary AI models handle abstract reasoning tasks when trained under conditions of incomplete rule exposure. Both vision-based and sequence-based models achieve high accuracy on familiar rules yet struggle to generalize when encountering omitted or novel ones. The sequence-to-sequence transformer consistently outperformed vision models, but its sharp performance decline in unseen-rule scenarios indicates that its reasoning remains fragile and heavily tied to training distributions. Furthermore, the discrepancy between token-level accuracy and correct-answer accuracy illustrates that high surface-level performance can mask deeper reasoning failures. These results suggest that current architectures often rely on memorization and local pattern matching rather than true analogical reasoning. Future work should focus on developing models that integrate symbolic structure with neural representations, incorporate stronger inductive biases for relational reasoning, and evaluate under conditions that demand extrapolation beyond the training space. 

\section*{Acknowledgments}
The author would like to thank the UCLA Cognitive Vision and Computational Learning (CVCL) Lab for providing support and resources throughout this work. Special thanks to Shuhao Fu for insightful guidance and helpful discussions that greatly contributed to the development of this study.

\bibliographystyle{unsrt}  
\bibliography{references}  


\end{document}